\documentclass[a4paper,conference]{IEEEtran}
\IEEEoverridecommandlockouts
\usepackage{cite}
\usepackage{subfigure}
\usepackage{caption}
\usepackage{subcaption}
\usepackage{amsmath,amssymb,amsfonts}
\usepackage{algorithmic}
\usepackage{graphicx}
\usepackage{textcomp}
\usepackage{xcolor}
\usepackage{multirow}
\usepackage{array}
\usepackage{rotating}
\usepackage{float}
\usepackage[section]{placeins}
\usepackage{hyperref}
\hypersetup{
  colorlinks   = true, 
  urlcolor     = blue, 
  linkcolor    = blue, 
  citecolor   = green 
}
\usepackage{tabularx,ragged2e,booktabs,caption}
\def\BibTeX{{\rm B\kern-.05em{\sc i\kern-.025em b}\kern-.08em
    T\kern-.1667em\lower.7ex\hbox{E}\kern-.125emX}}
    
    \makeatletter
\newcommand{\linebreakand}{%
  \end{@IEEEauthorhalign}
  \hfill\mbox{}\par
  \mbox{}\hfill\begin{@IEEEauthorhalign}
}
\makeatother

\begin{document}

\title{ Saliency-Guided Deep Learning for Bridge Defect Detection in Drone Imagery\\

}

\author{\IEEEauthorblockN{Loucif Hebbache }
\IEEEauthorblockA{\textit{dept. of Computer Science and Engineering} \\
\textit{Université du Québec en Outaouais}\\
Gatineau, Canada \\
}
\and
\IEEEauthorblockN{Dariush Amirkhani}
\IEEEauthorblockA{\textit{dept. of Computer Science and Engineering} \\
\textit{Université du Québec en Outaouais}\\
Gatineau, Canada \\
}
\linebreakand 
\IEEEauthorblockN{Mohand Saïd Allili}
\IEEEauthorblockA{\textit{dept. of Computer Science and Engineering} \\
\textit{Université du Québec en Outaouais}\\
Gatineau, Canada \\
}
\and 
\IEEEauthorblockN{Jean-François Lapointe}
\IEEEauthorblockA{\textit{Digital Technologies Research Centre} \\
\textit{National Research Council Canada}\\
Ottawa, Canada \\
}
}

\maketitle

\begin{abstract}
Anomaly object detection and classification are one of the main challenging tasks in  computer vision and pattern recognition. In this paper, we propose a new method to automatically detect, localize and classify defects in concrete bridge structures using drone imagery. This framework is constituted of two main stages. The first stage uses saliency for defect region proposals where  defects often exhibit local discontinuities in the normal surface patterns with regard to their surrounding. The second stage employs a YOLOX-based deep learning detector that operates on saliency-enhanced images obtained by applying bounding-box level brightness augmentation to salient defect regions. Experimental results on standard datasets confirm the performance of our framework and its suitability in terms of accuracy and computational efficiency, which give a huge potential to be implemented in a self-powered inspection system.

\end{abstract}

\begin{IEEEkeywords}
Bridge Defect Detection, Deep Learning, Saliency, Visual inspection, YOLOX.
\end{IEEEkeywords}

\section{Introduction}
Inspection to detect surface defects is an important task for maintaining the structural reliability of bridges. Failing to do so can lead to disastrous consequences as shown by the recent collapse of the Morandi bridge \cite{Calvi2019}. According to public data, out of 607,380 existing bridges in USA, nearly 67,000 are classified as structurally deficient whereas approximately 85,000 are considered functionally obsolete. Currently, the inspection task is often conducted manually by inspectors which could be time-consuming and sometimes cumbersome and painstaking. Recently, there has been a growing shift toward using aerial vehicles for performing inspection tasks, specifically for bridges, due to the provided benefits \cite{Kim2018}. Drones can reduce dramatically inspection time which is important to detect early stage damages to prevent any further losses in their structural capacity and durability. Using drones allow also to avoid dangerous situations for professional inspectors. They can basically collect visual data in hard-to-reach sites, but also autonomously perform real-time analysis of the images to assess about potential defects. Beyond basic imaging, recent AI–AR systems have shown that UAV-based bridge inspection can be carried out remotely with real-time defect overlays and interactive visualization, further reducing on-site time and cognitive load for inspectors \cite{Lapointe2022a,Lapointe2022b}.

Recent developments in deep learning (DL), specifically for object detection and classification, have unleashed huge opportunities for applications such as monitoring and visual inspection for anomaly detection and robust object detection in adverse imaging conditions \cite{Li2021,Bouafia2025,Mundt2019}. This also led to publishing of several benchmark datasets with annotations, facilitating model training and testing. One of the most popular datasets is  COncrete DEfect BRidge IMage dataset (CODEBRIM) proposed by Mundt et al. \cite{Mundt2019},  which exhibits multiple defects, including : crack, spallation, exposed reinforcement bar, efflorescence, corrosion (stains). This gave rise to several works that used machine learning techniques to detect bridge surface anomalies \cite{Feroz2021} and, more generally, to support visual maintenance of other civil infrastructure components such as sidewalks \cite{Park2025}.. Comprehensive surveys such as \cite{Amirkhani2024} provide an up-to-date overview of visual concrete bridge defect classification and detection using deep learning, highlighting the need for robust multi-defect and multi-scale models that operate reliably on UAV imagery.

For example, authors in \cite{Kim2018} used CNNs for detecting cracks via image classification. This method has shown a good performance, but shadows and low contrast can affect its accuracy. In \cite{Zhang2019}, the authors have proposed to use a single-stage for defect detection and classification based on the \textit{You Only Look Once} V3 (YOLOv3) model \cite{Redmon2018}. The approach has succeeded in detecting multiple damages in concrete bridges by achieving 80\%\ accuracy on the CODEBRIM dataset and out-performing the two-stage R-CNN method \cite{Ren2017}. However, it has shown some limitations when dealing with small and low-contrasted defects. In the same vein, Tabernik et al.\cite{Tabernik2020} used segmentation-based deep learning architecture to specifically tackle crack detection. More recently, Amirkhani et al. proposed CrackSight, an efficient crack segmentation model that maintains high accuracy across varying acquisition ranges and complex backgrounds, further improving robustness for real inspection scenarios \cite{Amirkhani2025}. Finally, \cite{Jiang2021} used YOLO  and MobileNetV3-SSD object detection algorithms to detect and classify concrete defects, which showed very promising results. More recent work has started to exploit the latest YOLO family members for concrete and building defect detection. For instance, Tian et al.\ proposed YOLOv11-EMC, an optimized YOLOv11-based framework with deformable and dynamic convolutions that improves precision, recall, and F1 for multi-category concrete surface defects, including drone field tests \cite{Tian2025}. At a broader level, Sapkota et al.\ reviewed the evolution of YOLO models from YOLOv1 up to the recently unveiled YOLOv12, showing consistent progress in speed and accuracy for real-time object detection and outlining future directions for YOLOv11/YOLOv12-based applications in infrastructure monitoring \cite{Sapkota2024}.

All the above methods have achieved noticeable successes when tackling individual defect classes such as cracks, but they loose effeciency when dealing with other types of defects. Indeed, developing a unique model that can deal with all defect classes hard to achieve. Defect detection and classification, and particularly on bridge surfaces, pose several challenges. First, even if defects can be categorized into different classes, there is huge intra-class variability due to varying illumination conditions, angle of view changes and size and severity of defects \cite{Feroz2021}. On the other hand, some defect classes can have a huge overlapping, while co-occurrence of different defects at some locations is not uncommon. Finally, datasets often contain noisy labels which add an extra difficulty in correctly training and evaluating algorithms \cite{Natarajan2013}. Saliency information has also been leveraged more explicitly in recent one-stage multi-label detectors for concrete defects, where Grad-CAM-based saliency maps are fused into a single-stage network to emphasize defect regions and improve performance on CODEBRIM and UAV datasets \cite{Hebbache2023}.

In this paper, we propose a method for automatic detection, localisation, and classification of defects using drone imagery. Our method is composed of two main steps. In the first step, we compute local saliency maps to highlight potential defect regions and derive bounding boxes around them, and then apply bounding-box level brightness augmentation to these salient regions to generate a saliency-enhanced image. In the second step, we use a multi-label YOLOX model \cite{Ge2021} operating on the saliency-enhanced image to localize and classify defects. Finally, we prune over- and under-detections by analysing the extent and continuity of the detected defects. Our method has been evaluated on the CODEBRIM \cite{Mundt2019} dataset, and the results show that it outperforms other state-of-the-art methods. 

This paper is organized as follows: Section \ref{sec1} presents the proposed method, Section  \ref{sec2} presents some experimental results for validation. We end the paper with a conclusion and future work perspectives.

\section{Proposed method}
\label{sec1}
\begin {figure*}[!htb]
    \centering
   \includegraphics[width=\textwidth]{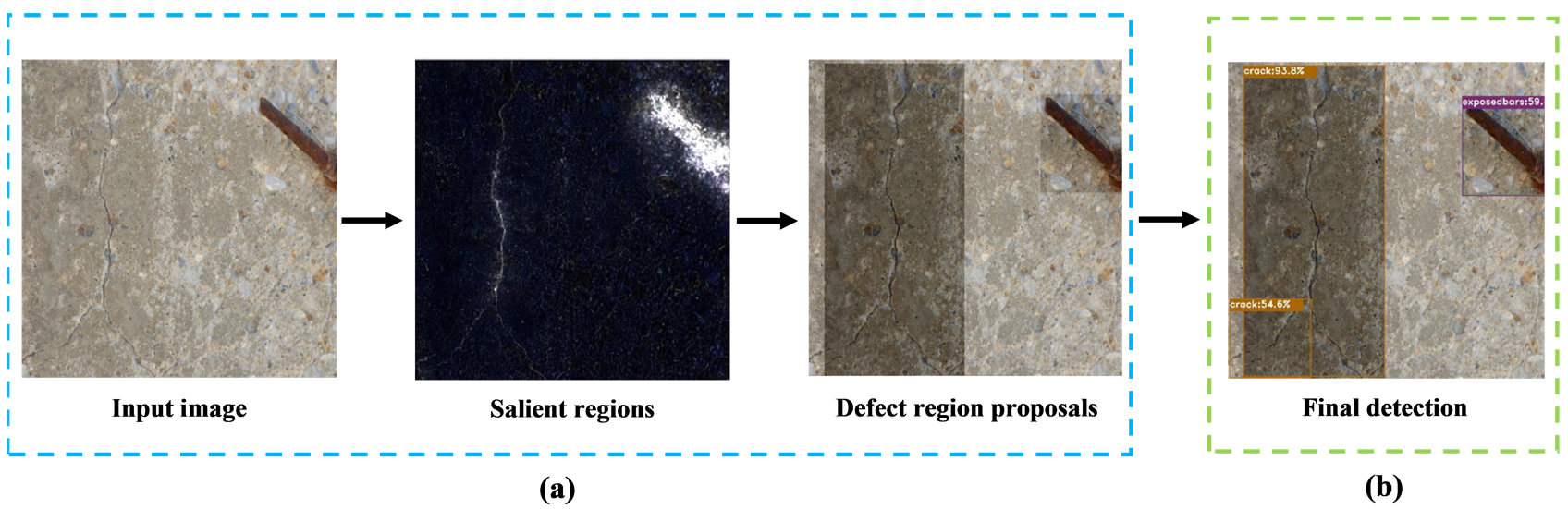}
    \caption{Our proposed method. (a): Saliency
and region proposal module; (b): Deep learning defect detection module.}
    \label{fig1}
\end{figure*}

Our core idea in this paper is to improve defect detection and classification by leveraging saliency as a front-end enhancement module for a deep object detector. This is motivated by the observation that, regardless of its type, a defect is typically characterized by a local surface discontinuity that can be more effectively captured by analysing local saliency than by relying solely on conventional object detection. Consequently, integrating saliency with a deep detector in a cascaded manner is particularly appealing for boosting defect detection performance, as it allows us to emphasize likely defect regions before performing detection. For this purpose, we design the pipeline depicted in Fig.~\ref{fig1}. 

Our method is composed of two main modules: a) a saliency-based defect region enhancement module, and b) a deep learning defect detection module. In the first stage, a visual saliency algorithm is used to highlight potential defect regions and to derive bounding boxes around these regions. Bounding-box level brightness augmentation is then applied to the salient regions, yielding a saliency-enhanced image in which defects are visually emphasized. In the second stage, a YOLOX model, fine-tuned using the CODEBRIM dataset, operates directly on the saliency-enhanced image to localize and classify defects. In the following, we describe each module in detail.

\subsection{Saliency for defect region proposals}
Saliency enables to identify regions that exhibit local discontinuities with regard to their surrounding. While not every salient region constitute a defect, most defects exhibit some degree of saliency. Thus, computing  saliency and use it to guide detection can be very useful for reducing false negatives which is a common problem in multi-class defect detection \cite{Jeong2020}. Note also that we tested different saliency methods, and the best result was obtained by the SmoothGrad \cite{Smilkov2017} which
extracts noisy sensitivity maps representing  most often key parts of the image. This is obtained by averaging maps made from several small perturbations of a given image followed by an enhancement.

The key idea behind the SmoothGrad method is to identify regions of an image that influence the most activation shifts in classification outputs.
Suppose $S_c$ is the class activation function. A mathematically way of locating “important” pixels (or \emph{sensitivity map}) of the input image
is to differentiate the activation function $S_c$ with respect to the input image $x$. The sensitivity map is given as follows:
\begin{equation}
M(x) = {\partial S_c(x)}/{\partial x}
\end{equation}

We observed that SmoothGrad generate very good defect locations especially when the defect occupies a large  surface. However, it performed poorly on highlighting linear structures that characterize some defects such as  cracks, and efflorescence at a smaller extent. To address this limitation, we build another map dedicated to linear pattern extraction from the image. The linearity map is generated by applying morphological bottom-hat filter $F$ to the image \cite{Gonzalez2018} and then subtracting the original image from the filtered one. This map is finally added to SmoothGrad ones to obtain a high quality map highlighting most of the defects in the image. 
\begin{equation}
D(x) = M(x) + |x- F(x)|
\end{equation}

To introduce the extracted salient regions into the subsequent detection stage, we first derive bounding boxes around the most salient areas and then apply brightness augmentation to the pixels within these boxes. This operation produces a saliency-enhanced image in which potential defect regions are locally brightened with respect to their surroundings. The original image structure is preserved outside these regions. Fig.~\ref{fig1}.(a) shows an example where all defects are highlighted in the saliency map. We can observe that the crack on the left, for instance, is clearly emphasized even though its contrast is not significant in the original image.

\subsection{Deep learning defect detection}

The issue of defect detection is not only to classify defects but also to localize them inside bounding boxes. Note that one-stage detection models are mainly focused on computation efficiency, which enables fast object detection at the expense of a limited accuracy \cite{Tan2019}. Here, we try to gain the benefits of the two worlds by simply narrowing  the search area of the one-stage detector, thus enabling rapid and accurate defect detection at once. Using the computed saliency maps, we generate region candidates around salient areas and apply bounding-box level brightness augmentation to these regions, producing saliency-enhanced images. The YOLOX detector then operates on these enhanced images, both during training and inference. In this way, saliency acts as a front-end transformation that increases the prominence of potential defect regions before they are processed by the one-stage detector, which is conceptually related to the notion of focusing attention on informative regions in two-stage object detectors \cite{Liu2020}.

For most of object detection problems, the bounding boxes are assigned one label. In our defect detection problem, this is not always verified since several defects can co-occur together.  For example, \emph{exposed bar} is often linked to \emph{corrosion} or \emph{spalling} defects. Likewise, \emph{cracks} can be linked to \emph{efflorescence} defects. Given this reality, we chose to fine-tune a YOLOX model with multi-label setting enabling to assign more than one label for each bounding box. The YOLOX model is the latest version of YOLO models, improving the limitation in terms of speed and accuracy. Compared to the one-stage detectors in the YOLO family,  YOLOX maintains the most important advantages including the removal of box anchors (increase the portability of the model to edge devices) and separating the YOLO detection head into feature channels for box classification and box regression (increase training convergence time and model accuracy). Fig. \ref{fig1}.(b) shows an illustration on how the YOLOX performs on the defect detection problem. We can clearly note that the defects have been well detected.\\

\section{Experiments and results}
\label{sec2}
To evaluate our method,  we conducted experiments on bridge defect detection and compared our results with previous methods. Here, we  briefly present the  dataset and evaluation metrics used 
for evaluation and well as some important implementation details about our method.

The most recent dataset that exists for bridge defecter detection is  CODEBRIM (COncrete DEfect BRidge IMage Dataset) \cite{Mundt2019} witch is composed of six classes: Background (2490) and five defect classes : \emph{Crack} (2507), \emph{Spallation} (1898), \emph{ Exposed bar} (1507), \emph{Efflorescence} (833), and \emph{Corrosion} (1559). The images were acquired at high-resolution using drone, then resized to fit the input resolution required by YOLOX. In order to make the model perform better, especially on small objects and with changing viewing conditions, we augmented this dataset using both image and bounding box level augmentation techniques such as : brightness, mosaic and shear. Bounding box level augmentation generates new training data by only altering the content of a source images bounding boxes \cite{Zoph2020}. \\ 

For our the experiments, we used Google Colab Pro+, which provides 52GB of RAM alongside 8 CPU cores and priority access to GPU P100. The CODEBRIM dataset is split into training, validation, and test sets with an approximate ratio of 70\%, 20\%, and 10\%. We trained YOLOX along 100 epochs with 1182 iterations for each epoch, a 8 batch size and fp16 mixed-precision training. To help reduce False Positives (FP), we add 10\% background images to the dataset, with no objects(labels). When using inference with YOLOX, to select the best bounding box from the multiple predicted bounding boxes, we use the Non-Maximal Suppression (NMS) technique, which is included in YOLOX with a threshold equal to 0.45.

To evaluate our method performance and compare it to other models, we use the mAP (Mean Average Precision) metric representing the mean value of the AP of all classes. The AP is the area under the precision-recall curve. The precision and recall metrics are defined as follows:

\begin{equation}
Recall = \frac{TP}{TP + FN}
\end{equation}

\begin{equation}
Precision = \frac{TP}{TP + FP}
\end{equation}

Where TP, FP, and FN represent the number of true positives, false positives, and false negatives, respectively.To classify the detections as TP or FP, the IOU (Intersection Over Union) threshold is set to t = 0.5 and t = 0.75.

\subsection{Results Analysis}

To see the benefit of using saliency, we first conducted object detection experiments directly using YOLOX without using saliency. The obtained AP values for each  as well as the  mAP value are presented in Table \ref{tab1}. 

\begin{table}[h]
\caption{Results of model evaluation on test set.}
    \centering
    \begin{tabular}{c|c|c|c}

    \textbf{Classes} & \textbf{AP@0.5} & \textbf{mAP@0.5} & \textbf{mAP@0.95}\\
    \hline \hline
    \centering \arraybackslash Crack & 0.91 &  \multirow{5}{*}{\textbf{0.91}} & \multirow{5}{*}{\textbf{0.87}}\\
    Spallation & 0.91 & &\\
    Efflorescence & 0.91 & &\\
    Exposedbars & 0.91 & &\\
    Corrosionstain & 0.90 & &\\
    \hline \hline

    \end{tabular}
    \label{tab1}
\end{table}

\begin{figure}[!htb]
\begin{subfigure}
\centering
\includegraphics[width=.45\linewidth]{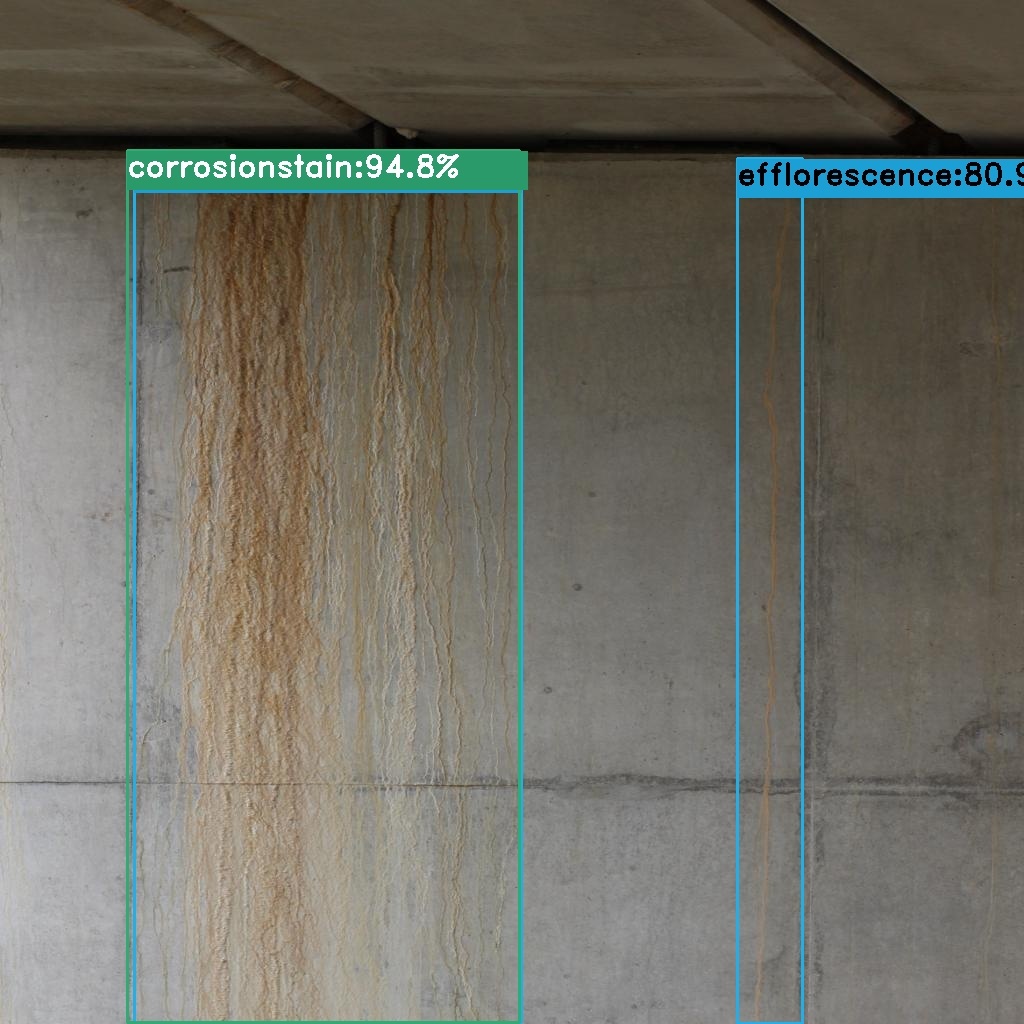}
\includegraphics[width=.45\linewidth]{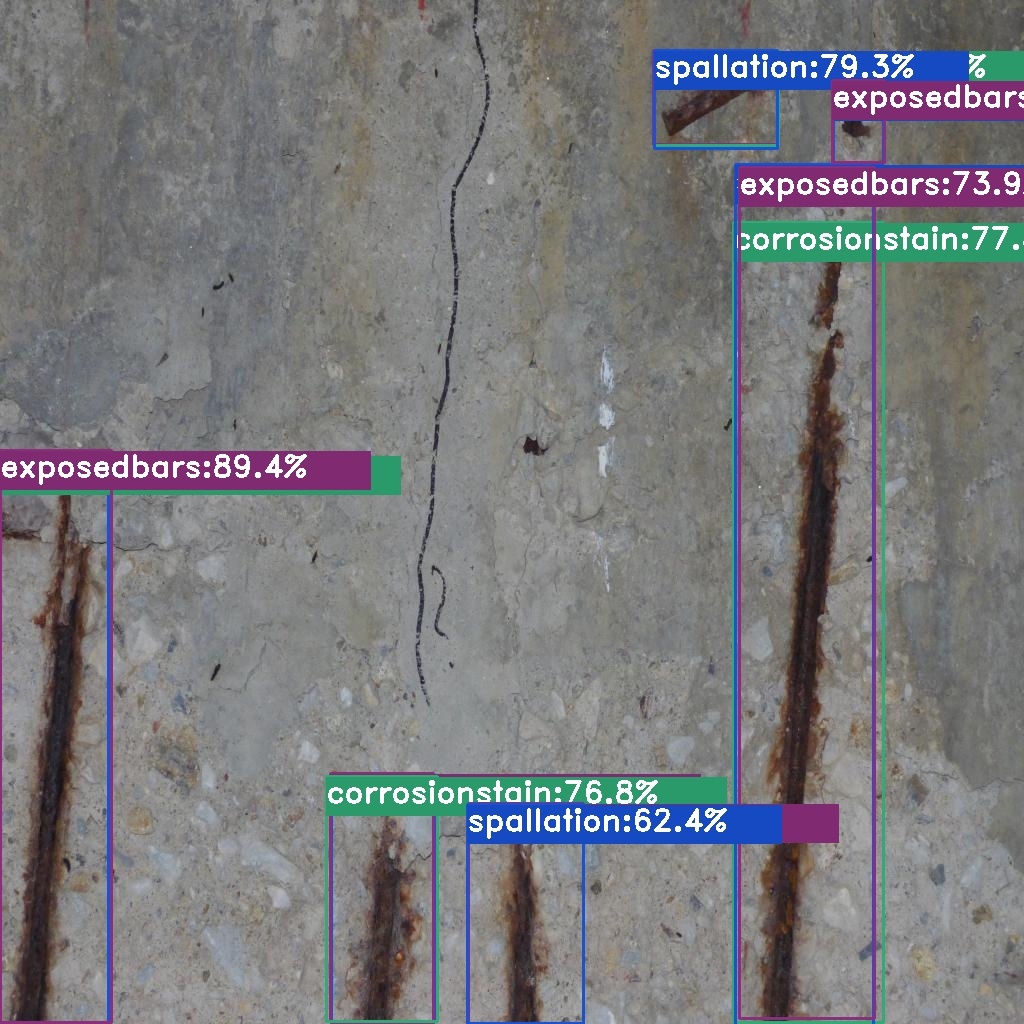}
\includegraphics[width=.45\linewidth]{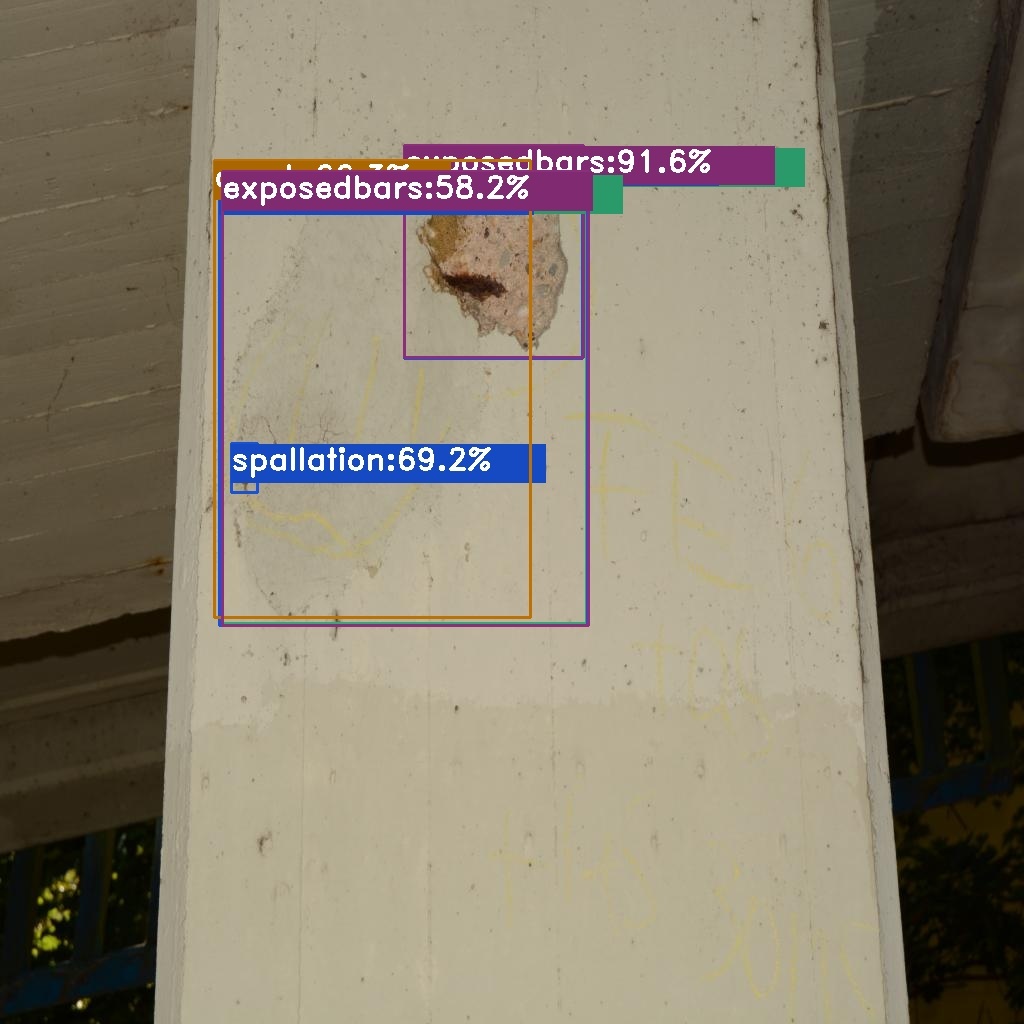}
\includegraphics[width=.45\linewidth]{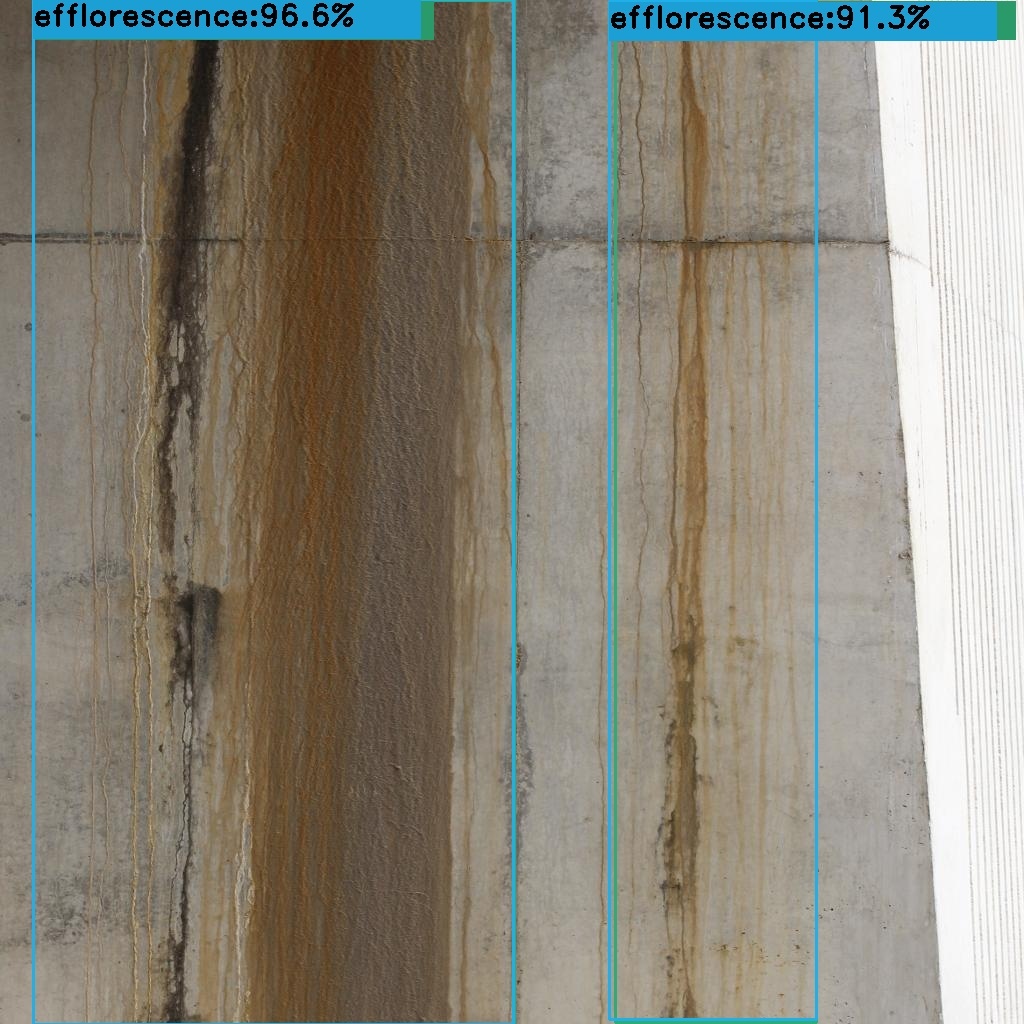}
\caption{Detection samples by YOLOX : defects well detected.}
\label{fig2}
\end{subfigure}
\end{figure}

\begin{figure}[!htb]

\begin{subfigure}

\centering
\includegraphics[width=.45\linewidth]{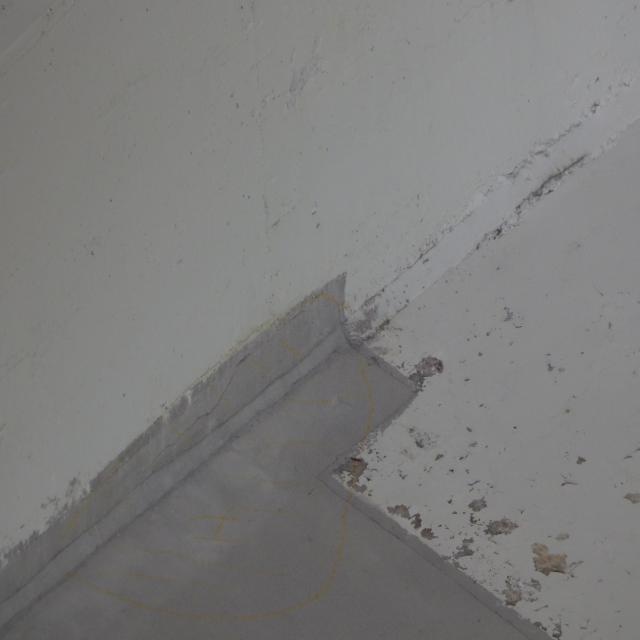}
\includegraphics[width=.45\linewidth]{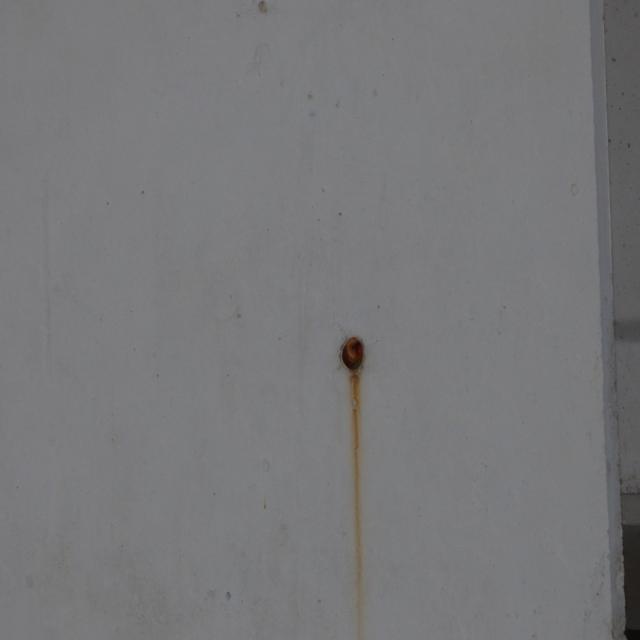}
\includegraphics[width=.45\linewidth]{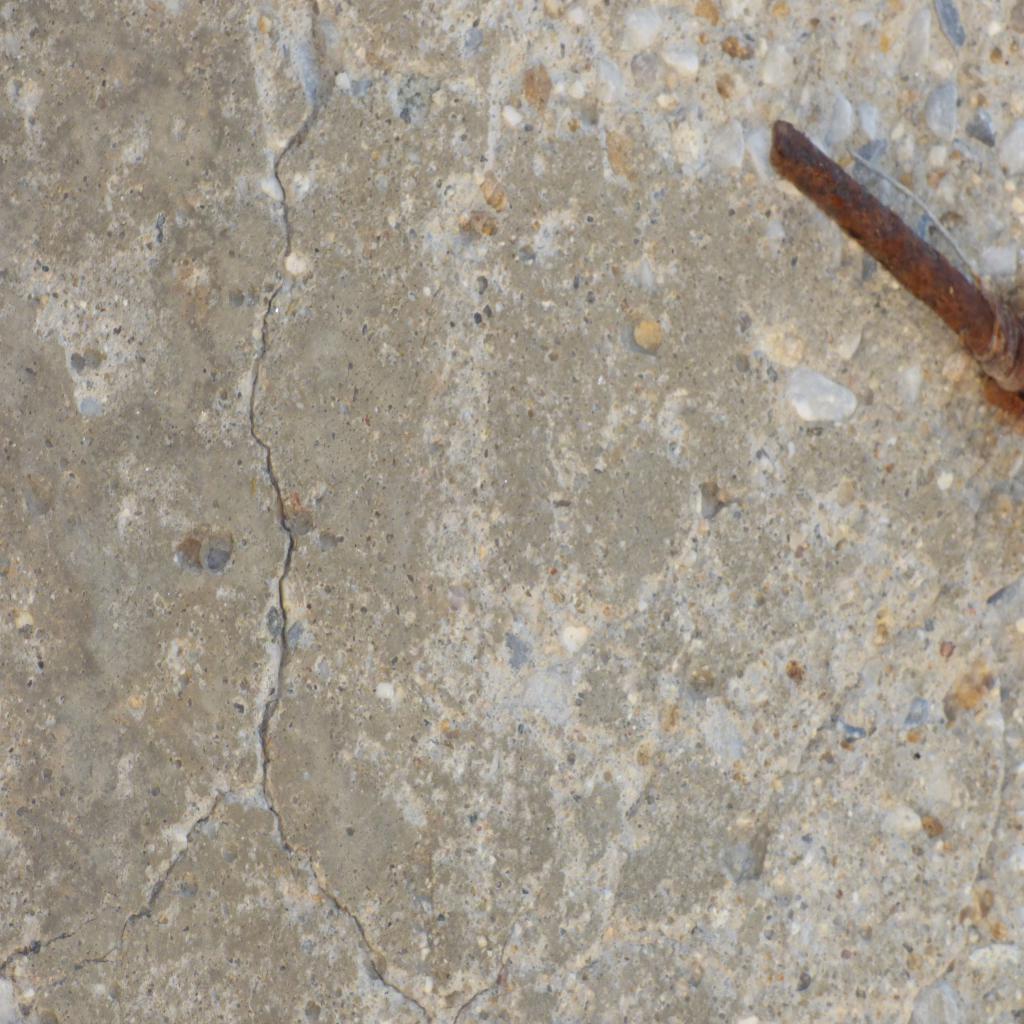}
\includegraphics[width=.45\linewidth]{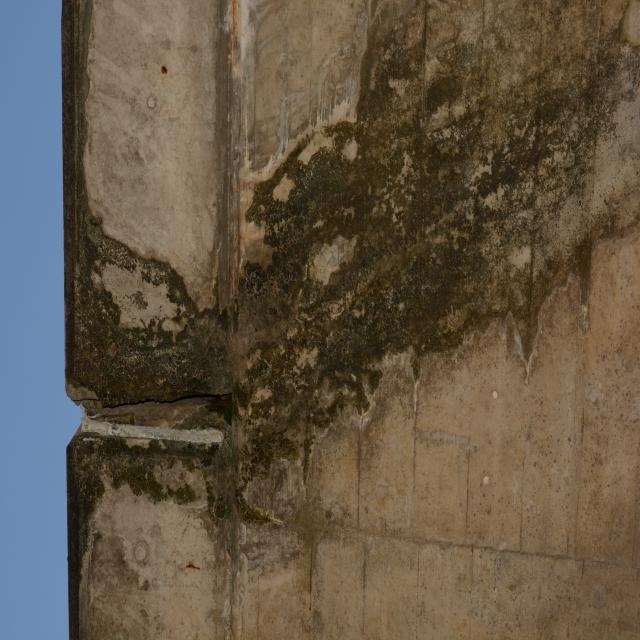}

\caption{Detection samples by YOLOX : missed defects.}
\label{fig3}
\end{subfigure}

\end{figure}

\begin{figure}[!htb]

\begin{subfigure}

\centering
\includegraphics[width=.45\linewidth]{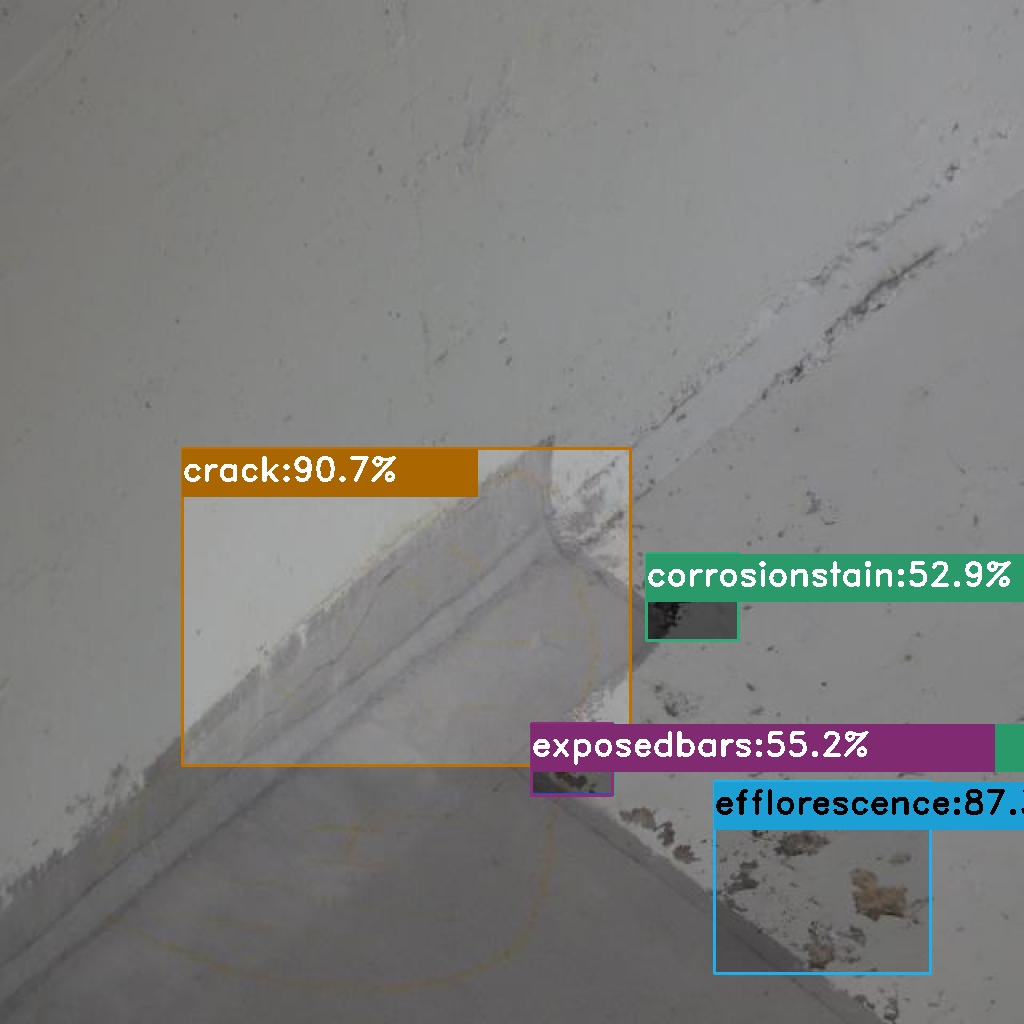}
\includegraphics[width=.45\linewidth]{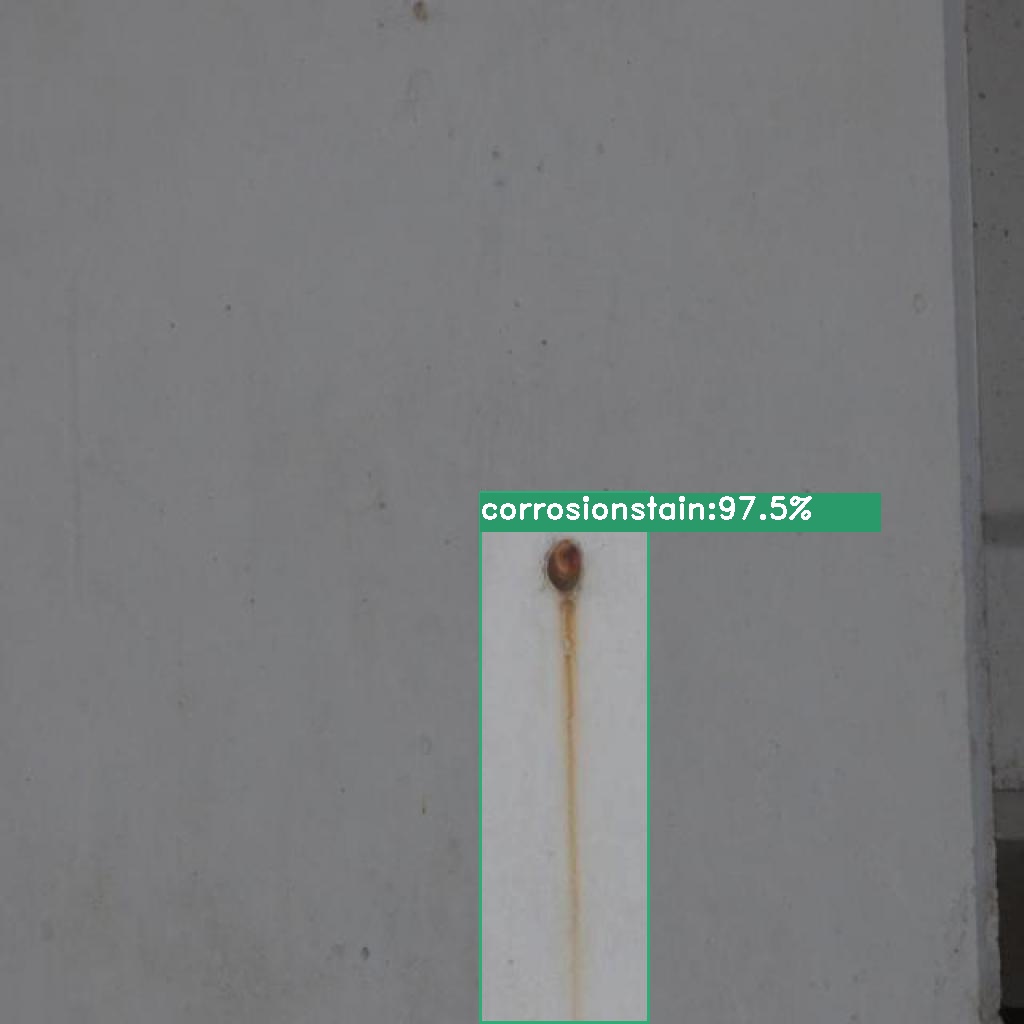}
\includegraphics[width=.45\linewidth]{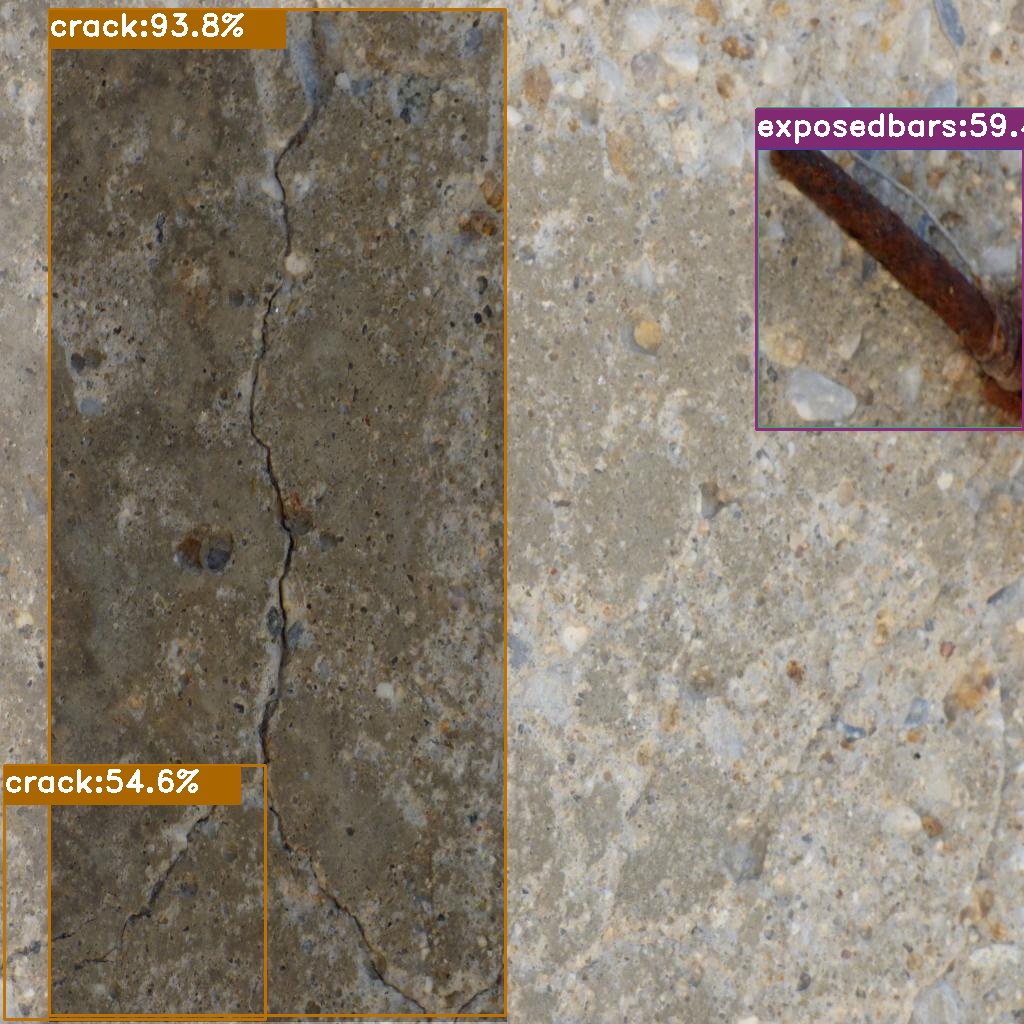}
\includegraphics[width=.45\linewidth]{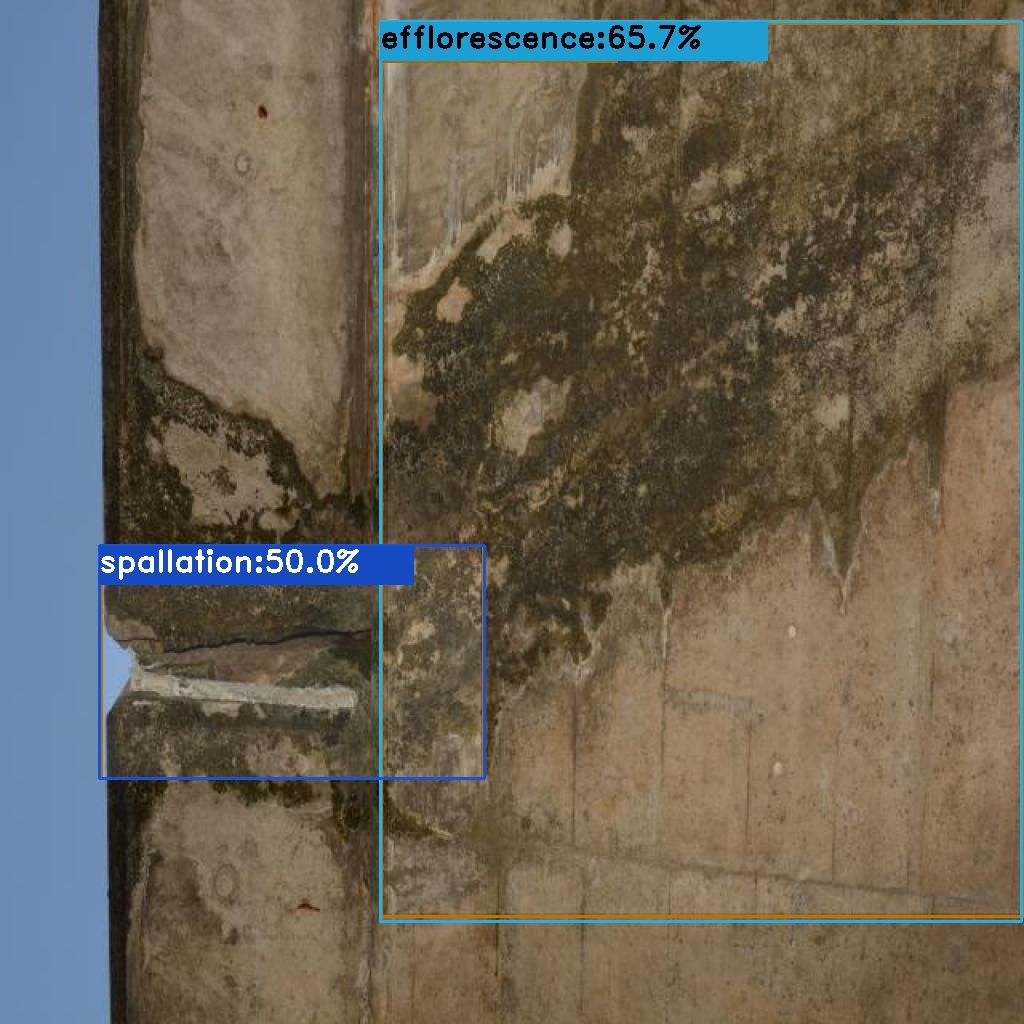}

\caption{Results detection of the previous missed samples with our method.}
\label{fig4}
\end{subfigure}

\end{figure}

Fig. \ref{fig2} presents  some detections samples where defects are well detected by YOLOX. Fig. \ref{fig3} shows some detection samples where YOLOX completely missed defects. We observed that the missed defects are often very small or do not present sufficient contrast with regard to the background.  Consequently, YOLOX failed to detect them. Our method using saliency has enabled the detection of almost all the missed defects. Table \ref{tab2} shows defect detection results of YOLOX after applying saliency compared without saliency. As we can see, YOLOX with saliency performed better than YOLOX. Fig. \ref{fig4} shows qualitative  results of defect detection of the previous missed samples with our method.

The proposed method was compared against several recent object detectors, including Improved YOLOv3 and SSD from Jiang et al. \cite{Jiang2021}, Improved YOLOv3 and Faster R-CNN from Zhang et al. \cite{Zhang2020}, as well as RetinaNet \cite{Lin2017}, YOLOv5-l \cite{Jocher2020}, YOLOv8-l \cite{Jocher2023}, YOLOR-P6 \cite{Wang2021}, and SMDD-Net \cite{Hebbache2023}. Table~\ref{tab:comparison} summarizes the comparison results. Our saliency-guided YOLOX achieves the highest mAP@0.5, slightly surpassing SMDD-Net and substantially outperforming the other detectors.  

\begin{table}[!htb]
\caption{Results of YOLOX with saliency and without saliency.}
    \centering 
    \begin{tabular}{c|c|c|c}
    
    \textbf{Method} & \textbf{mAP@0.5} & \textbf{mAP@0.75} & \textbf{mAP@0.95} \\
    \hline \hline
    \centering YOLOX without saliency & 0.91 &  0.89 & 0.87 \\
     \centering YOLOX with saliency & \textbf{0.99} & \textbf{0.97} &  \textbf{0.95} \\
    \hline \hline
    \end{tabular}
    \label{tab2}
\end{table}

\begin{table}[!htb]
\caption{Comparison of our method with previous studies on CODEBRIM (mAP@0.5).}
\centering
\begin{tabular}{l|c}
\textbf{Method} & \textbf{mAP@0.5 (\%)} \\
\hline\hline
Jiang et al. (YOLOv3) \cite{Jiang2021}      & 64.8 \\
Jiang et al. (SSD) \cite{Jiang2021}         & 64.1 \\
Zhang et al. (YOLOv3) \cite{Zhang2020}      & 79.9 \\
Zhang et al. (Faster R-CNN) \cite{Zhang2020}& 74.4 \\
YOLOv5-l \cite{Jocher2020}                  & 41.7 \\
YOLOv8-l \cite{Jocher2023}                  & 59.6 \\
RetinaNet \cite{Lin2017}              & 88.4 \\
YOLOR-P6 \cite{Wang2021}                   & 89.2 \\
SMDD-Net \cite{Hebbache2023}                 & 99.1 \\
\textbf{Ours (Saliency-Guided YOLOX)}       & \textbf{99.6} \\
\hline\hline
\end{tabular}
\label{tab:comparison}
\end{table}


\section{Conclusion}
\label{sec3}
In this paper, we developed a saliency-guided deep learning method for bridge defect detection based on YOLOX and drone imagery. The proposed approach comprises two cascaded stages: first, visual saliency is computed to highlight potential defect regions and to derive bounding boxes around them; second, bounding-box level brightness augmentation is applied to these salient regions to produce a saliency-enhanced image, which is then processed by a YOLOX detector to localize and classify defects. Experiments on the CODEBRIM dataset show that the proposed method improves defect detection accuracy by 8.3\% compared with using YOLOX alone. Moreover, quantitative comparisons with a range of state-of-the-art detectors, including YOLOv5, YOLOv8, RetinaNet, YOLOR-P6, and SMDD-Net, demonstrate that our approach achieves the highest mAP@0.5. As future work, we plan to integrate saliency more tightly into the YOLOX architecture, for example by incorporating saliency-based attention mechanisms directly within the detection network.

\section{Acknowledgements}
This project was supported in part by collaborative research funding from the National Research Council of Canada’s Artificial Intelligence for Logistics Program. Thanks also to the NRC's National Program Office for its support.


\begin{thebibliography}{1}
\bibitem{Amirkhani2025}
D. Amirkhani, M. S. Allili, and J. F. Lapointe. CrackSight: An efficient crack segmentation model in varying acquisition ranges and complex backgrounds. IEEE Transactions on Automation Science and Engineering, 2025 Jul 21. (Early Access)

\bibitem{Amirkhani2024}
D. Amirkhani, M. S. Allili, L. Hebbache, N. Hammouche, and J. F. Lapointe. Visual concrete bridge defect classification and detection using deep learning: A systematic review. IEEE Transactions on Intelligent Transportation Systems, vol. 25, no. 9, pp. 10483–10505, 2024.

\bibitem{Baker2016}
B. Baker, O. Gupta, N. Naik, and R. Raskar, ,Designing neural network architectures using reinforcement learning. arXiv preprint arXiv:1611.02167, 2016.

\bibitem{Bouafia2025}
Y. Bouafia, M. S. Allili, L. Hebbache, and L. Guezouli. SES-ReNet: Lightweight deep learning model for human detection in hazy weather conditions. Signal Processing: Image Communication, vol. 130, p. 117223, 2025.

\bibitem{Bozic2020}
J. Bozic, D. Tabernik and D. Skocaj. End-to-end training of a two-stage neural network for defect detection. \emph{Int'l Conf. On pattern Recognition}, 5619-5626, 2020.

\bibitem{Bukhsh2021}
Z.A. Bukhsh, N. Jansen, and A. Saeed. Damage detection using in-domain and cross-domain transfer learning. arXiv preprint arXiv:2102.03858.2021

\bibitem{Calvi2019}
G.M. Calvi et al., Once upon a time in Italy: The tale of the morandi bridge. \textit{Structural Engineering Int'l}, vol. 29, no. 2, pp. 198–217, 2019

\bibitem{Czimmermann2020}
T. Czimmermann et al., Visual-based defect detection and classification approaches for industrial applications—a survey. \textit{Sensors}, 20(5), p. 1459, 2020.

\bibitem{Feroz2021}
 S. Feroz and S. Abu Dabous. UAV-Based Remote Sensing Applications for Bridge Condition Assessment. \textit{Remote Sensing}, 13, 1809, 2021.

 \bibitem{Feng2019}
C. Feng, H. Zhang, S. Wang, Y. Li, H. Wang, and F. Yan. Structural
damage detection using deep convolutional neural network and transfer
learning. KSCE Journal of Civil Engineering, vol. 23, no. 10, pp. 4493– 4502, 2019.


\bibitem{Ge2021}
Z. Ge, S. Liu, F. Wang, Z. Li, and J. Sun, “Yolox: Exceeding yolo series
in 2021,” arXiv preprint arXiv:2107.08430, 2021.

\bibitem{Gonzalez2018}
R.C. Gonzalez and R.E. Woods. Digital Image Processing, Pearson 4th Edition, 2018.
\bibitem{Hu2021}
Hu, J.Y., Shi, C.J.R. and Zhang, J.S., 2021. Saliency-based YOLO for single target detection. Knowledge and Information Systems, 63(3):717-732.

\bibitem{Hebbache2023}
L. Hebbache, D. Amirkhani, M. S. Allili, N. Hammouche, and J. F. Lapointe. Leveraging saliency in single-stage multi-label concrete defect detection using unmanned aerial vehicle imagery. Remote Sensing, vol. 15, no. 5, p. 1218, 2023.

\bibitem{Huthwohl2019}
P. Hüthwohl, R. Lu,  and I. Brilakis. Multi-classifier for reinforced concrete bridge defects. Automation in Construction, 105, p.102824,2019

\bibitem{Jiang2021}
Y. Jiang, D. Pang, and C. Li.A deep learning approach for fast detection and classification of concrete damage. Automation in Construction, 128, p. 103785, 2021.

\bibitem{Jeong2020}
E. Jeong, J., Seo and J. Wacker. Literature Review and Technical Survey on Bridge Inspection Using Unmanned Aerial Vehicles. \textit{J. of Performance of Constructed Facilities}, 34: 04020113, 2020

\bibitem{Kim2018}
I.-H. Kim et al., Application of Crack Identification Techniques for an Aging Concrete Bridge Inspection Using an Unmanned Aerial Vehicle. \emph{Sensors}, 18(6), 1881, 2018.
\bibitem{Lapointe2022a}
J. F. Lapointe, M. S. Allili, L. Belliveau, L. Hebbache, D. Amirkhani, and H. Sekkati. AI-AR for bridge inspection by drone. In International Conference on Human-Computer Interaction (HCII 2022), May 29–June 3, 2022, ser. Lecture Notes in Computer Science, pp. 302–313. Cham: Springer International Publishing, 2022.

\bibitem{Lapointe2022b}
N. Hammouche, J. F. Lapointe, H. Sekkati, M. S. Allili, L. Hebbache, and D. Amirkhani. AI-AR for remote visual bridge inspection by drone. In 11th International Conference on Structural Health Monitoring of Intelligent Infrastructure (SHMII-11), 2022, pp. 462–465.

\bibitem{Jocher2020}
G. Jocher. YOLOv5. Online: \url{https://github.com/ultralytics/yolov5}, accessed Feb. 27, 2023.

\bibitem{Jocher2023}
G. Jocher. YOLOv8. Online: \url{https://github.com/ultralytics/ultralytics}, accessed Feb. 27, 2023.

\bibitem{Lin2017}
T.-Y. Lin, P. Goyal, R. Girshick, K. He, and P. Doll\'ar, ``Focal loss for dense object detection,'' in \emph{Proc. IEEE Int. Conf. Comput. Vis. (ICCV)}, 2017, pp. 2980--2988.

\bibitem{Wang2021}
C.-Y. Wang, I.-H. Yeh, and H.-Y. M. Liao, ``You only learn one representation: Unified network for multiple tasks,'' arXiv preprint arXiv:2105.04206, 2021.

\bibitem{Li2021}
C-L. Li et al., CutPaste: Self-Supervised Learning for Anomaly Detection and Localization.  \emph{IEEE Conf. on Computer Vision and Pattern Recognition}, 9664-9674, 2021.

\bibitem{Liu2020}
L. Liu et al., Deep Learning for Generic Object Detection: A Survey. \textit{Int'l J. of Computer Vision}, 128:261-318, 2020.

\bibitem{Mundt2019}
M. Mundt et al., Meta-learning convolutional neural architectures for multi-target concrete defect classification with the concrete defect bridge image dataset.
\textit{IEEE Conference on Computer Vision and Pattern Recognition}, 11196–11205, 2019.

\bibitem{Natarajan2013}
N. Natarajan et la., Learning with Noisy Labels. \emph{Advances in Neural Information Processing Systems} , 2013.

\bibitem{Redmon2018}
J. Redmon and  A. Farhadi. YOLOv3: An Incremental Improvement, arXiv:1804.02767v1 2018.

\bibitem{Park2025}
J. Park, S. Lee, P. H. Cho, Z. Xia, and L. Hebbache. Enhancing sidewalk maintenance through accurate joint deflection measurement. In 2025 IEEE International Conference on Consumer Electronics (ICCE), Jan. 11–13, 2025, pp. 1–4. IEEE, 2025.

\bibitem{Pham2018}
H. Pham, M. Guan, B. Zoph, Q. Le, and J. Dean. Efficient neural architecture search via parameters sharing. Intl'l Conf. on Machine Learning, 4095–4104, 2018.

\bibitem{Rakha2018}
R.T. Gorodetsky. A Review of Unmanned Aerial System (UAS) applications in the built environment: Towards automated building inspection procedures using drones. \textit{Automation in Construction}, 93: 252–264, 2018.

\bibitem{Ren2017}
S. Ren, K. He, R. Girshick and J. Sun. Faster R-CNN: Towards Real-Time Object Detection with Region Proposal Networks. \emph{IEEE Trans. on Pattern Analysis and Machine Intelligence},
39: 1137-1149, 2017.

\bibitem{Smilkov2017}
D. Smilkov et al., Smoothgrad: removing noise by adding noise. arXiv preprint arXiv:1706.03825, 2017.

\bibitem{Sapkota2024}
R. Sapkota, R. Qureshi, M. F. Calero, C. Badjugar, U. Nepal, A. Poulose, P. Zeno, U. B. Vaddevolu, S. Khan, M. Shoman, and H. Yan. YOLOv12 to its genesis: A decadal and comprehensive review of the You Only Look Once (YOLO) series. arXiv preprint arXiv:2406.19407, 2024.

\bibitem{Tan2019}
Z. Tan et al. Learning to Rank Proposals for Object Detection. \emph{IEEE Int'l Conf on Computer Vision},  8272-8280, 2019.

\bibitem{Tian2025}
Z. Tian, F. Yang, L. Yang, Y. Wu, J. Chen, and P. Qian. An optimized YOLOv11 framework for the efficient multi-category defect detection of concrete surface. Sensors, vol. 25, no. 5, p. 1291, 2025.

\bibitem{Tabernik2020}
D. Tabernik et al., Segmentation-based deep-learning approach for surface-defect detection. Journal of Intelligent Manufacturing, 31(3):759–776, 2020.

\bibitem{Zhang2019}
C. Zhang, C.-C. Chang and M. Jamshidi. Concrete bridge surface damage detection using a single‐stage detector. \emph{Computer-Aided Civil and Infrastructure Engineering}, 35(9) 389-409, 2019.

\bibitem{Zhang2020}
C. Zhang, C.-c. Chang, and M. Jamshidi. Concrete bridge surface damage detection using a single-stage detector. Computer-Aided Civil
and Infrastructure Engineering, vol. 35, no. 4, pp. 389–409, 2020.
\bibitem{Zoph2020}
B. Zoph , ED. Cubuk, G. Ghiasi, TY. Lin, J. Shlens, and QV. Le.  Learning data augmentation strategies for object detection. InEuropean Conference on Computer Vision ,  (pp. 566-583), 2020.



\end{thebibliography}

\end{document}